%% file: main.tex
\documentclass[10pt,twocolumn,letterpaper]{article}

\usepackage{iccv}              % To produce the CAMERA-READY version
\usepackage{booktabs} % For formal tables
\usepackage{enumitem}
\usepackage{graphicx, animate}
\usepackage{caption}
\usepackage{subcaption}

\input{preamble}

\definecolor{iccvblue}{rgb}{0.21,0.49,0.74}
\usepackage[pagebackref,breaklinks,colorlinks,allcolors=iccvblue]{hyperref}

\title{MultiCOIN: Multi-Modal COntrollable Video INbetweening}

\author{%
    Maham Tanveer$^{1,2}\thanks{Work done as intern at Adobe.}$ \quad Yang Zhou$^2$ \quad Simon Niklaus$^2$ \\
    Ali Mahdavi Amiri$^1$ \quad Hao Zhang$^1$ \quad Krishna Kumar Singh$^2$ \quad Nanxuan Zhao$^2$ \\
    $^1$Simon Fraser University \qquad $^2$Adobe Research\\ 
}

\begin{document}
%\maketitle

\input{figs/teaser.tex}

\maketitle

\maketitle

\input{sec/0_abstract}

\input{sec/1_intro}

\input{sec/2_related}

\input{sec/3_method}
\input{sec/4_results}
\input{sec/5_conc}
% adding inputs here 
\clearpage
{
    \small
    \bibliographystyle{ieeenat_fullname}
    \bibliography{main}
}
\end{document}

%% file: preamble.tex
\usepackage{xcolor}
\usepackage{pifont}
\usepackage{booktabs} % For formal tables
\usepackage{enumitem}
\usepackage{graphicx, animate}
\usepackage{caption}
\usepackage{subcaption}
\usepackage{amssymb}     % for checkmark
\usepackage[normalem]{ulem} % in your preamble

\definecolor{darkgreen}{rgb}{0.0, 0.5, 0.0}

\newcommand{\cmark}{\textcolor{darkgreen}{\checkmark}}
\definecolor{turquoise}{rgb}{0.0,0.5,0.5}
\newcommand{\xmark}{\textcolor{red}{\ding{55}}}    % red cross
\usepackage{pifont}      % for ding symbols
\usepackage{float}
\usepackage{dblfloatfix} % allows better placement of figure*
\usepackage{multirow}

\newcommand{\modelname}[1]{\noindent{MultiCOIN}}

%% file: figs/teaser.tex
% \teaser{
%  \includegraphics[width=0.9\linewidth]{data/teaser_new.pdf}
%  \centering
%   \caption{
%     Our model, \modelname{}, takes a start and end image frame to generate an interpolative video {\em inbetweening\/}. It supports {\em multi-modal\/} controls, including depth change and layering, motion trajectories, text prompts, and target regions for movement localization, to generate smooth and plausible transitions.
%     %
%     % For example, depth can be used to enable diverse motion control (top two rows). Guide pixels and text can be used together or independently. Simple trajectory control is also supported (third-row). These conditions may be mixed and matched.
%     The control can be used individually (top four rows) to create diverse results even with the same input pair (e.g., different depth layering results in top two rows). The control can also be organized in a general complementary way to ease the user's interactions. For example, target regions may be used for content control, while trajectory provides motion information. Also, while specifying the general movement of the woman by text, the user can exert accurate spatial control for the bird with target region.
%     }
% \label{fig:teaser}
% }

\twocolumn[{
\maketitle
\begin{center}
    \centering
    \includegraphics[width=1\textwidth]{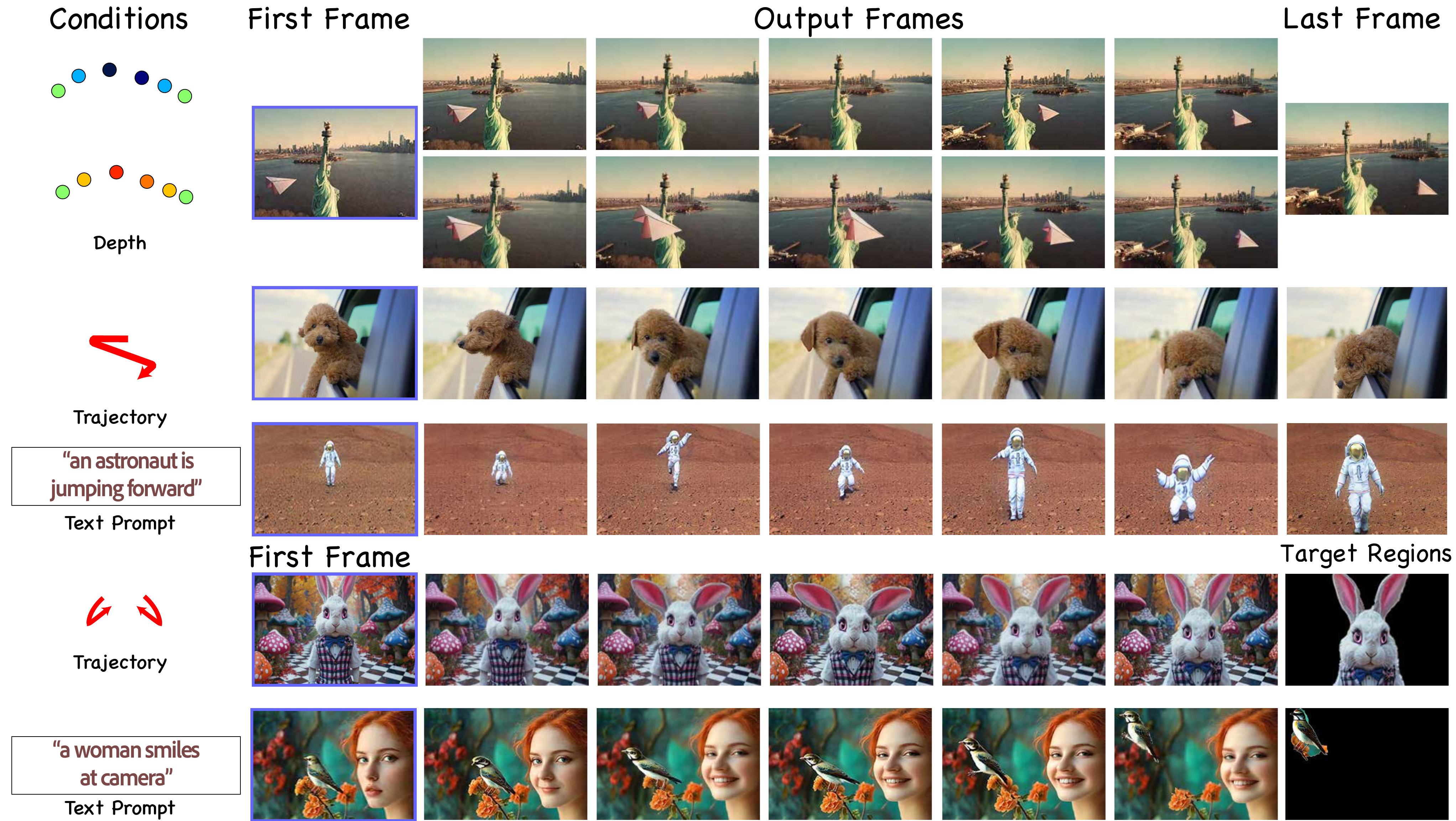}
    \captionof{figure}{
    Our model, \modelname{}, takes a start and end image frame to generate an interpolative video {\em inbetweening\/}. It supports {\em multi-modal\/} controls, including depth change and layering, motion trajectories, text prompts, and target regions for movement localization, to generate smooth and plausible transitions.
    %
    % For example, depth can be used to enable diverse motion control (top two rows). Guide pixels and text can be used together or independently. Simple trajectory control is also supported (third-row). These conditions may be mixed and matched.
    The control can be used individually (top four rows) to create diverse results even with the same input pair (e.g., different depth layering results in top two rows). The control can also be organized in a general complementary way to ease the user's interactions. For example, target regions may be used for content control, while trajectory provides motion information. Also, while specifying the general movement of the woman by text, the user can exert accurate spatial control for the bird with target region.
    }
    \label{fig:teaser}
\end{center}%
}]

%% file: sec/0_abstract.tex
\begin{abstract}
Video inbetweening creates smooth and natural transitions between two image frames, making it an indispensable tool for video editing and long-form video synthesis. Existing works in this domain are unable to generate large, complex, or intricate motions. In particular, they cannot accommodate the versatility of user intents and generally lack fine control over the details of intermediate frames, leading to misalignment with the creative mind. 
To fill these gaps, we introduce \modelname{}, a video inbetweening framework that allows multi-modal controls, including depth transition and layering, motion trajectories, text prompts, and target regions for movement localization, while achieving a balance between flexibility, ease of use, and precision for fine-grained
video interpolation. To achieve this, we adopt the Diffusion Transformer (DiT) architecture as our video generative model, due to its proven capability to generate high-quality long videos. To ensure compatibility between DiT and our multi-modal controls, we 
map all motion controls into a common sparse and user-friendly point-based representation as the video/noise input.
Further, to respect the variety of controls which operate at
varying levels of granularity and influence, we separate content controls and motion controls into two branches to encode the required features before guiding the denoising process, resulting in two generators, one for motion and the other for content. Finally, we propose a stage-wise training strategy to ensure that our model learns the multi-modal controls smoothly. 
Extensive qualitative and quantitative experiments demonstrate that multi-modal controls enable a more dynamic, customizable, and contextually accurate visual narrative.

\href{https://multicoinx.github.io/multicoin/}{Project Page: MultiCOIN}

\end{abstract}

% We design generators to extract the control signals faithfully and encode features through dual-branch embedders to resolve ambiguities and seamlessly integrate the motion and content controls in generating natural-looking video outputs. We further introduce a stage-wise training strategy to smoothly learn various controls. Extensive qualitative and quantitative experiments demonstrate that multi-modal controls enable a more dynamic, customizable, and contextually accurate visual narrative.

%-------------------------------------------------------------------------
%  ACM CCS 1998
%  (see https://www.acm.org/publications/computing-classification-system/1998)
% \begin{classification} % according to https://www.acm.org/publications/computing-classification-system/1998
% \CCScat{Computer Graphics}{I.3.3}{Picture/Image Generation}{Line and curve generation}
% \end{classification}
%-------------------------------------------------------------------------
%  ACM CCS 2012
   % (see https://www.acm.org/publications/class-2012)
%The tool at \url{http://dl.acm.org/ccs.cfm} can be used to generate
% CCS codes.
%Example:

% \begin{CCSXML}
% <ccs2012>
%    <concept>
%        <concept_id>10010147.10010178.10010224.10010240.10010241</concept_id>
%        <concept_desc>Computing methodologies~Image representations</concept_desc>
%        <concept_significance>500</concept_significance>
%        </concept>
%  </ccs2012>
% \end{CCSXML}

% \ccsdesc[500]{Computing methodologies~Image representations}

% \printccsdesc

%% file: sec/1_intro.tex
\section{Introduction}
\label{sec:intro}

Video inbetweening or video interpolation seeks to generate intermediate frames between two end keyframes, creating a smooth transition from one scene to another. It is a long-standing problem~\cite{choi2000fruc, ha2004motion} and an increasingly important building block for video content creators and animators as they perform video editing, storytelling, and short-to-long video synthesis~\cite{meyer2018colorprop, siyao2021animeinterp}. Such a frame interpolation is typically carried out in two steps: motion estimation and motion compensation~\cite{ha2004moco, choi2000fruc, niklaus2020softsplat, reda2022film}. 
As temporal and spatial gaps between the input frames widen, both tasks are faced with
significant challenges, since generating realistic intermediate frames requires synthesizing novel content to fill in and bridge the missing information, as well as resolving the inherent ambiguities therein. However, as the emerging generative models~\cite{blattmann2023align} become more powerful, the continuing growth of the space of exploration for the generated frames has opened up new possibilities for inbetweening of distant input scenes. 
At the same time, this poses a one-to-many problem, where a single output is typically insufficient since users are often not interested in just {\em any\/} possible video interpolation, but one which respects their artistic expression or creative mind. As a result, the inbetweening must be {\em user-controllable\/}.
%as can be reflected via a variety of controls.}

Prior attempts on controllable video inbetweening, as exemplified by the recent work Framer~\cite{wang2024framer}, have focused on respecting motion trajectories. In practice however, user controls are often more versatile and fine-grained. Recent advances in LLMs have popularized the use of text prompts as a means for edit controls. Even when confined to trajectory guidance only, additional constraints such as {\em depth transitions\/} (e.g., to specify whether an object moves in front of or behind another object, as shown in the top row of Fig.~\ref{fig:teaser}) and {\em region/object localization\/} (e.g., see the bottom two rows of Fig.~\ref{fig:teaser} for the use of {\em target regions\/} to isolate the motioned object) must be incorporated to avoid ambiguities.

In this paper, we present \modelname{}, for {\em MULTI-modal COntrollable INbetweening\/}, a novel video inbetweening framework which can accommodate all the edit controls mentioned above, as shown in Fig.~\ref{fig:teaser}. Specifically, trajectory-based controls provide precise motion paths. 
Depth inputs can add 3D structure cues to help disambiguate non-lateral motions and improve occlusion handling. Furthermore, target regions can add motion localization constraints, ensuring consistency over detailed, especially multi-object, regions, while text-based control facilitates high-level semantic guidance. By combining all of these modalities, our method strives to achieve a balance between flexibility, ease of use, and precision, empowering users to achieve high-quality and fine-grained video interpolation with minimal effort.

To generate dynamic, accurate, and customizable motion transitions with multiple controls, we must build on an advanced video generative model. To this end, we resort to
the Diffusion Transformer (DiT) architecture~\cite{peebles2023scalable}, due to its proven capability to generate high-quality long videos, which our method targets. 
%Indeed, controllability is especially critical when editing long videos with the end frames being far apart. 
The first challenge however, lies in making the multi-modal controls compatible with DiT. Unlike UNet, adopted by Framer for trajectory control only, DiT employs a Vision Transformer (ViT)-style 3D Variational Autoencoder (VAE) that divides frames into spatio-temporal patches with positional encodings and compresses them along the temporal dimension. These operations disrupt the spatial correlation of native control signals, e.g., for trajectory and depth, making them incompatible in their raw forms. Likewise, content information provided at different temporal locations must be aligned with DiT’s representation space as well. 

To resolve the incompatibility, we map all the controls into the same domain as the video/noise input. First, trajectory and depth information, presented in the form of optical flow and depth maps respectively, are converted into RGB, as the VAE in DiT operates on such a format.
%using visualization methods; 
Specifically, depth is represented using relative color encoding and applied to both compositional layering (see Fig. \ref{fig:teaser}, top two rows) and for object-specific control (see Fig. \ref{fig:sparse_depth}).
Next, we transform dense optical flow and depth maps into {\em sparse\/}, point-based representations by extracting trajectories from high-motion regions. Along these trajectories, both optical flow and depth values are sampled, yielding sparse control points, which are more user-friendly. Users may provide one or both of these modalities, trajectory and depth, depending on the desired level of control. 
%Such a design choice makes the controls user-friendly: instead of requiring full maps, a user can specify sparse points to guide generation. 

The input frames, including those defining target regions (see Fig.~\ref{fig:teaser}, bottom), are inserted at designated temporal positions with the remaining slots filled with black frames and corresponding binary masks indicating valid regions. These representations are passed through the DiT-VAE and appended to the DiT input noise.

The second challenge arises when we must generate intermediate frames coherently with the keyframes in both spatial and temporal domains, while respecting the variety of controls which operate at varying levels of granularity and influence. To this end, we separate content controls, given by the input frames, from motion controls, via optical flow and depth, into {\em two branches\/} to encode the required features before guiding the denoising process, resulting in two generators, one for motion and the other for content. Our experiments have demonstrated that such a dual-branch setup provides greater stability and robustness in training, while preserving both motion fidelity and content consistency, despite the multi-modality. Finally, we propose a stage-wise training strategy to ensure that our model learns the multi-modal controls smoothly. Specifically, we feed the model with denser and more concrete controls first, and then gradually move to sparser and higher-level controls.

We evaluate \modelname{} through extensive quantitative and qualitative experiments. Our method supports multi-object control using trajectory and depth, with content guided by keyframes and target regions at different temporal points. Depth enables layering and object-specific control, while text prompts refine motion or act as standalone signals. By aligning motion controls with the input in the spatio-temporal domain, our approach achieves significantly better trajectory alignment compared to Framer. Moreover, \modelname{} demonstrates strong multi-modal versatility, highlighting the benefits of complementing trajectories with additional controls for more flexible inbetweening.

%% file: sec/2_related.tex
\input{figs/pipe.tex}
\section{Related Work}
\label{sec:related}
\subsection{Video Generation}

Creating realistic and novel videos has long been an interesting research problem~\cite{yu2023magvit,ranzato2016video}. Earlier studies have employed various generative models including GANs~\cite{yu2023magvit,saito2017temporal,tulyakov2017mocogan,shen2023mostganv} and temporally aware networks such as LSTM or autoregressive models~\cite{srivastava2016unsupervised,yan2021videogpt,hong2022cogvideo}. Recently, inspired by the success of diffusion models in image synthesis, several works have begun to investigate the use of diffusion models for conditional and unconditional video generation~\cite{ho2022video, singer2022make, ho2022imagen, kwon2024harivo}. Stable Video Diffusion~\cite{blattmann2023stable} leverages latent diffusion models~\cite{rombach2022high, blattmann2023align} for generating temporally coherent content. Few-shot video generation is facilitated by methods like Tune-a-video~\cite{wu2023tune}, which fine-tunes pre-trained image diffusion models, while training-free methods~\cite{hong2023large} leverage large language models for generative guidance. Another approach to generating videos in a controllable manner is to use keyframes along with text conditions~\cite{girdhar2023emu, wang2024microcinema, li2023videogen, zeng2024make}, where initial frames are generated to guide subsequent frames, with latent-consistency networks ensuring temporal and visual coherence. 
In this work, we target video inbetweening that aims to interpolate between two given frames following flexible multi-modal controls in a joint framework.
\subsection{Video Inbetweening}
Video inbetweening has a few other names such as frame interpolation, frame rate up-conversion, or temporal super-resolution. It has a long history, with early approaches operating at a block- instead of a pixel-level due to compute constraints at the time~\cite{choi2000fruc, ha2004moco}. 
While we have more compute nowadays, the underlying framework of motion estimation and compensation has largely remained the same throughout the years~\cite{niklaus2020softsplat, niklaus2023splatsyn, reda2022film, niklaus2018ctxsyn}. Flow-based methods use optical flow from the input frames for generating the inbetween frames  \cite{jiang2018super, park2020bmbc, huang2020rife}. While approaches like phase- or kernel-based interpolation~\cite{niklaus2017adaconv, niklaus2017sepconv, meyer2015phasebased,zhou2022audio} use spatially adaptive kernels to synthesize the interpolated pixels. In either case, it is fundamentally still about re-synthesizing an in-between frame from what is in the input frames. However, as the inputs become more distant in time and/or space, the inbetweening will require information that is not present so we need to hallucinate it instead. 
Nowadays we can utilize foundational video models for generating plausible interpolation results~\cite{feng2024explorative, danier2024ldmvfi, xing2025dynamicrafter}, but users typically aren't interested in just a possible interpolation result but one that follows their artistic expression. This is where motion control comes into the picture, which is the focus of our work. 
Framer~\cite{wang2024framer} is a work that achieves impressive results in controllable inbetweening using motion trajectories. Our method aligns trajectory control more effectively with the input in spatio-temporal domain, resulting in improved motion accuracy. In addition, it introduces a multi-modal framework that combines complementary controls to generate more diverse outputs. Qualitative comparisons in Fig.~\ref{fig:framer} highlight these improvements.
% \todoc{needs elaboration that why framer cannot be easily extended to other conditions.}
% \input{figs/main_results}

%% file: figs/pipe.tex
\begin{figure*}[t]
\centering
  \includegraphics[width=1\textwidth]{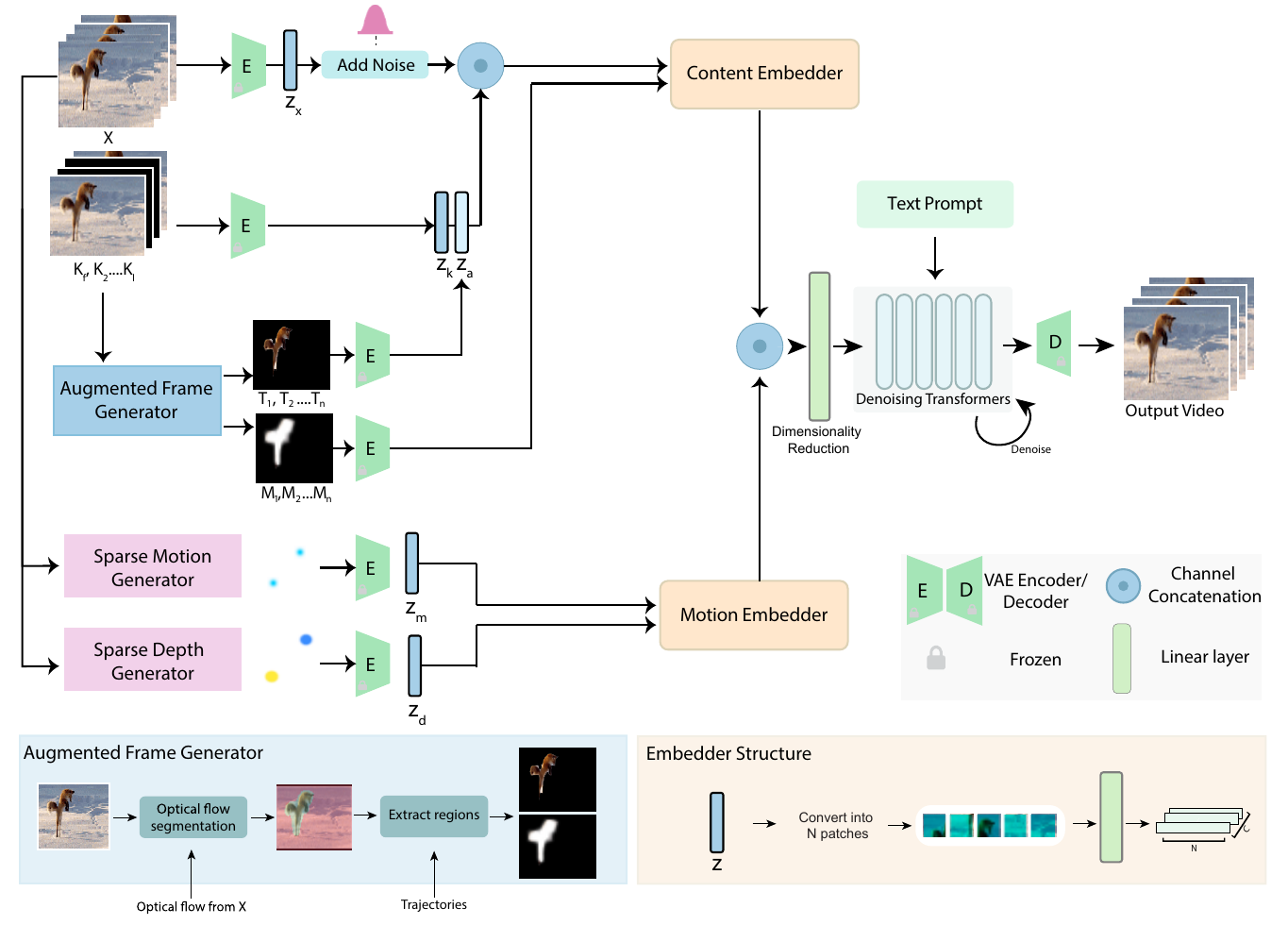}
  \caption{
  Overview of our \modelname{} pipeline. Given a video $X$, we extract multi-modal motion controls through two generators: the Sparse Motion Generator via optical flow and the Sparse Depth Generator for depth maps, both producing sparse RGB points for trajectory and depth control. An Augmented Frame Generator computes target regions and masks to enable fine-grained content control. All control signals are encoded via a dual-branch embedder architecture that separately captures motion and content features. In addition, a text prompt condition is processed by a text encoder to provide semantic guidance over the generated content. At inference, the model flexibly integrates these multi-modal controls for interpolation.
  }
  \label{fig:pipeline}
\end{figure*}

% Overview of our \modelname{} pipeline. Given a video $X$, we propose a  Sparse Motion Generator to provide conditioning for the motion trajectory with sparse RGB point controls, and an Augmented Frame Generator to compute guiding pixels for providing fine-grained control. The control signals are encoded through dual-branch embedders respectively to capture accurate content and motion features. Our model is flexible to take multi-modal controls for interpolation during the inference. 

%% file: sec/3_method.tex
\section{Method}
\label{sec:method}

Our goal with \modelname{} is to provide an intuitive and effective control mechanism for inbetweening tasks using motion (\eg, trajectories, depth) and content (\eg frames, target regions) guidance for intuitive and fine-grained control as shown in Fig.~\ref{fig:teaser} for generating realistic and coherent outputs. 
During the training, given the ground truth video clip $X$ with the extracted keyframes \(\{K_f, K_2, \ldots, K_l\}\), we represent the motion control as \textit{depth-trajectories} consisting of sparse points \(\{P_1, P_2, \ldots, P_m\}\) which have both directional and depth information. We use optical flow and depth maps visualized in RGB format, from which we then extract \(\{P_1, P_2, \ldots, P_m\}\) through the proposed \textit{Sparse Motion/Depth Generators}. Along with keyframes we add additional content control via target regions \(\{T_1, T_2, \ldots, T_n\}\) and associated guide masks \(\{M_1, M_2, \ldots, M_n\}\). These provide regional control in content generation. We extract them through the proposed \textit{Augmented Frame Generator}. 
Our overall model builds on a DiT-based video diffusion backbone, chosen for its ability to generate long and coherent videos. On top of this backbone, we thus propose two modules (1) Sparse Motion/Depth Generators, which produce the trajectory-based motion controls from RGB flow and depth maps, and (2) Augmented Frame Generator, which provides regional content controls through target regions and masks. These complementary modules are integrated through dual-branch embedders that separately encode motion and content controls. The following sections describe the DiT backbone and each proposed module.

\subsection{Preliminary}
Models like Stable Video Diffusion (SVD) are generative models that extend image diffusion to video by maintaining temporal consistency across frames. Given a noisy video \( X_T \), the model utilizes a conditional 3D-UNet to progressively denoise it to a clean video \( X_0 \) by iteratively applying a denoising function: $X_{t-1} = \epsilon_\theta(X_t, t, c)$, where \( \epsilon_\theta \) represents the learned noise, and $c$ represents conditions. Diffusion Transformer (DiT)~\cite{peebles2023scalable} models combine diffusion-based denoising processes with transformer architectures. Compared to traditional UNet-based models, DiT leverages a transformer backbone as its core denoiser to model long-range dependencies and global context, which is critical for capturing fine details and significantly improves the versatility and quality of image and video generation. For training, a diffusion loss is used which measures the mean square error (MSE) between the predicted noise $\hat{\epsilon}$ and the input noise $\epsilon$: $\mathcal{L}_{diff} = || \hat{\epsilon} - \epsilon ||^2_{2}.$

\input{figs/sparse_pipe}
\subsection{Control Generation}
\label{sec:control_gen}
Large-motion interpolation is challenging due to ambiguity, artifacts, and distortions, requiring precise and high-quality control.
Our method employs two mechanisms: 1) Sparse Motion-Depth Generator, which focuses on key motion paths, and 2) Augmented Frame Generator, which adds extra visual context. Together, these methods enhance the model’s ability to produce controlled, natural-looking motion.

\subsubsection{Sparse Motion-Depth Generators}
The Sparse Motion-Depth Generator (Fig.~\ref{fig:sparse_pipeline}) produces motion outputs aligned with both the model architecture and the input video $X$. A key challenge is generating motion inputs that are not only valid for our target scenario but also structurally compatible with the DiT framework. DiT uses a ViT-based 3D VAE to encode input videos by dividing frames into patches with learned positional encodings and compresses them along temporal dimension. Because our motion inputs are physically grounded (\eg optical flow and depth maps that directly control spatial displacements), ensuring their compatibility with this patch-based and temporally compressed representation is non-trivial.

First, in the absence of ground-truth motion trajectories, we generate them by extracting dense optical flow from the input video $X$ and tracking points with high motion magnitude. Along these tracked paths, we must extract both optical flow and depth information. 
To make these motion inputs compatible with DiT, we project all controls into the model’s latent space. We begin by converting both optical flow and depth trajectories into RGB representations. For optical flow, this involves mapping the direction and magnitude of motion to color space creating a visual representation.  For depth, relative distance values are mapped to a red–blue colormap, where hue encodes whether points lie inward or outward in the scene. These two visualizations enable the optical flow and depth inputs to be processed in the same way as RGB video frames.

We then extract sparse trajectories, as discussed above, and along these trajectories extract the optical flow and depth RGB values. Since each trajectory originates from a single pixel, the resulting motion signals are too sparse to be meaningful. To improve spatial coverage and interpretability, we expand each trajectory’s influence using 2D filters. For optical flow, we adopt a Gaussian filter, similar to~\cite{wang2024motionctrl}, which spreads motion smoothly while preserving directionality and gradually reducing magnitude. For depth, we instead use a disk filter that copies depth values over a circular region. Gaussian spreading is suited for optical flow, where hue encodes direction and gradual falloff naturally models motion attenuation, whereas depth hue directly encodes distance, so even small color changes correspond to different depth values. A uniform disk is therefore preferable to avoid introducing unintended depths. The sparse trajectories are thus converted into sparse RGB point controls ${P_1, P_2, \ldots, P_m} \in \mathbb{R}^{H \times W \times 3}$, following the same format as the input frames. These controls are then passed through DiT’s existing 3D VAE, allowing the model to effectively embed motion information, yielding promising results.

Another challenge with depth being a relative measure arises during inference, specifically for single-point inputs, where a depth reference is critical to anchor the model’s understanding. To address this, we compute the mean of the sparse depth values provided by the user along the trajectory and generate anchor points at three multiples above and below the mean. These anchors are placed at the corners of the depth input to supply global depth context, as illustrated in Fig.~\ref{fig:sparse_depth}.
\input{figs/sparse_depth}

\subsubsection{Augmented Frame Generator}
While motion paths provide effective control over inbetweening, we discovered that the inherent ambiguity of diffusion models, combined with the challenges of interpolating large motions, makes regional control an important enhancement. This approach refines the output, reducing the number of motion paths needed. At the same time, we want to avoid overly rigid control, allowing for more natural results. To achieve this, we introduce Augmented Frames. The core concept is to provide the model with a subtle ``nudge" in the right direction using content which we call ``Target Regions". This may be used alongside trajectories or on their own.

To implement this, we extract a region of interest from \(K_f\) and translate it across several frames according to the corresponding trajectory to create frames of target regions \(\{T_1, T_2, \ldots, T_n\}\), which are appended to \(K_f\) temporally. For training, we generate regions from motion trajectories using optical-flow segmentation. Further details are available in Fig.~\ref{fig:pipeline}. The ``fox" example in the figure illustrates how we extract the region corresponding to the direction of sparse motion trajectories and append it to the input keyframes. 

We also extract a binary mask that associates valid and invalid regions, and goes as an extra condition into content encoder. This helps the model to separate valid pixel information (in keyframes and target regions) from invalid information.
Once the model learns to interpret target regions, we can manually set these guiding regions. Users can specify exactly where the model should place content, such as moving a region from one spot to another. This allows explicit control over the generated frames. The target region control reduces the need for users to draw extensive trajectories, helping the model accurately identify and track the complete moving object with minimal input.
More details and results can be found in the experiment section.
During training, we dropout this content condition with probability of 50\%, making it optional at inference.

\input{figs/main_results}

% \subsection{Stage-wise Training for Multi-modal Control with Dual-Branch Encoders}
\subsection{Stage-wise Training with Dual-Branch Encoders}
\label{sec:cond_diff}
To train our model, we utilize a dual-branch encoder structure. First, a set of random keyframes \(\{K_f, K_2, \ldots, K_l\}\) is extracted from \(X\). First and last keyframe are always provided, and we select 0-5 random keyframes in between first and last keyframe to help the model learn multiple keyframe interpolation. We extract \(\{T_1, T_2, \dots, T_n\}\) from these keyframes, for which we use the dense optical flow of \(X\) to create sparse trajectories and optical flow segmentations. The first branch encodes the content information including  \(\{K_f, K_2, \ldots, K_l\}\), \(\{T_1, T_2, \dots, T_n\}\) and \(\{M_1, M_2, \dots, M_n\}\).

For motion, we extract \(\{P_1, P_2, \ldots, P_m\}\) which includes both motion and depth as discussed above and details in Fig~\ref{fig:sparse_pipeline}. The second branch encodes this motion information. Both branches have a similar structure. The input (motion or content) is first passed through a frozen VAE to encode it into a latent representation. For content, the latent representation of noise is channel-concatenated with the latent output of conditional images (keyframes and target regions). These latent outputs are then passed through Embedders, which first transform the inputs into patches and then funnels the output through a linear layer. These outputs are again channel-concatenated and passed through a final linear layer before being fed into the transformer denoiser.

To train our model, we utilize a stage-wise training strategy, where we gradually introduce conditional inputs to the model. First, the model is trained on the image branch alone to learn core video interpolation, ensuring it can interpolate between two images without conditions. Afterwards, to embed the motion and depth as a condition, we first performed a trial experiment. Using the architecture in Fig.~\ref{fig:pipeline} we directly train with \(\{P_1, P_2, \ldots, P_m\}\). From the results we saw that the model struggles to properly follow the motion and depth, specifically in localizing the movement. This analysis is discussed in Sec.~\ref{sec:abl_opt}. To address this issue, we adopted an alternative approach inspired by~\cite{yin2023dragnuwa, wang2024motionctrl}. 

We first trained the model solely with dense optical flow and dense depth maps, and then gradually introduced the sparse motion inputs. This phased approach enables the model to better interpret the limited motion information. In the last step, we train with guided pixels (\(\{T_1, T_2, \dots, T_n\}\)) and (\(\{M_1, M_2, \dots, M_n\}\)). Intuitively, we opted for a two-branch system to separate the two very different conditional inputs. In Sec.~\ref{sec:abl_opt} we show how this approach leads to better stability in the output.

%% file: figs/sparse_pipe.tex
\begin{figure}[h]
\centering
  \includegraphics[width=1\linewidth]{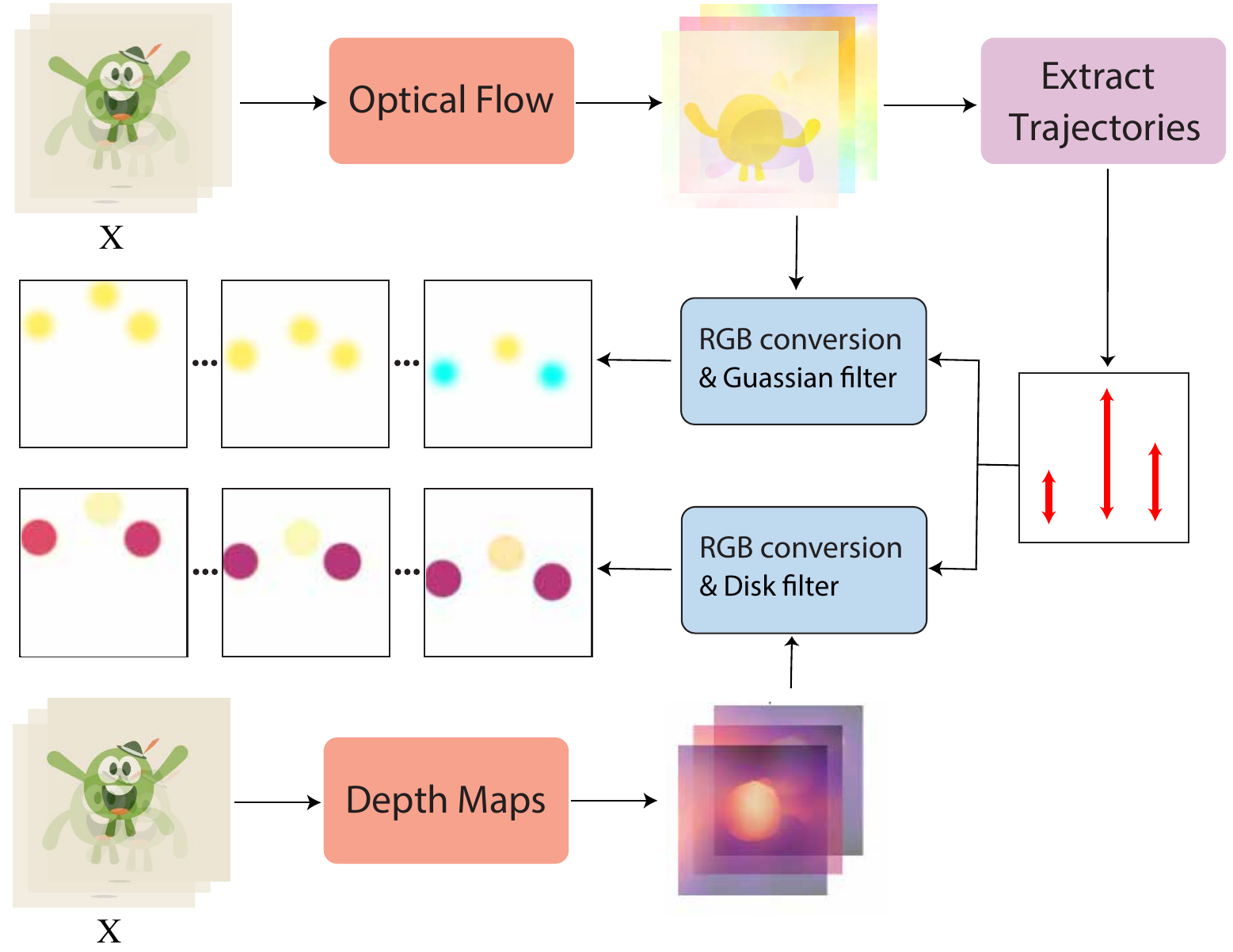}
  \caption{
  Sparse Motion and Depth Generator. Given video $X$, dense optical flow and depth maps are computed. Trajectories are selected from high-motion regions along which flow/depth points are sampled and expanded with 2D filters to get sparse RGB inputs.
  }
  \label{fig:sparse_pipeline}
\end{figure}

%% file: figs/sparse_depth.tex
\begin{figure}[h]
\centering
  \includegraphics[width=1\linewidth]{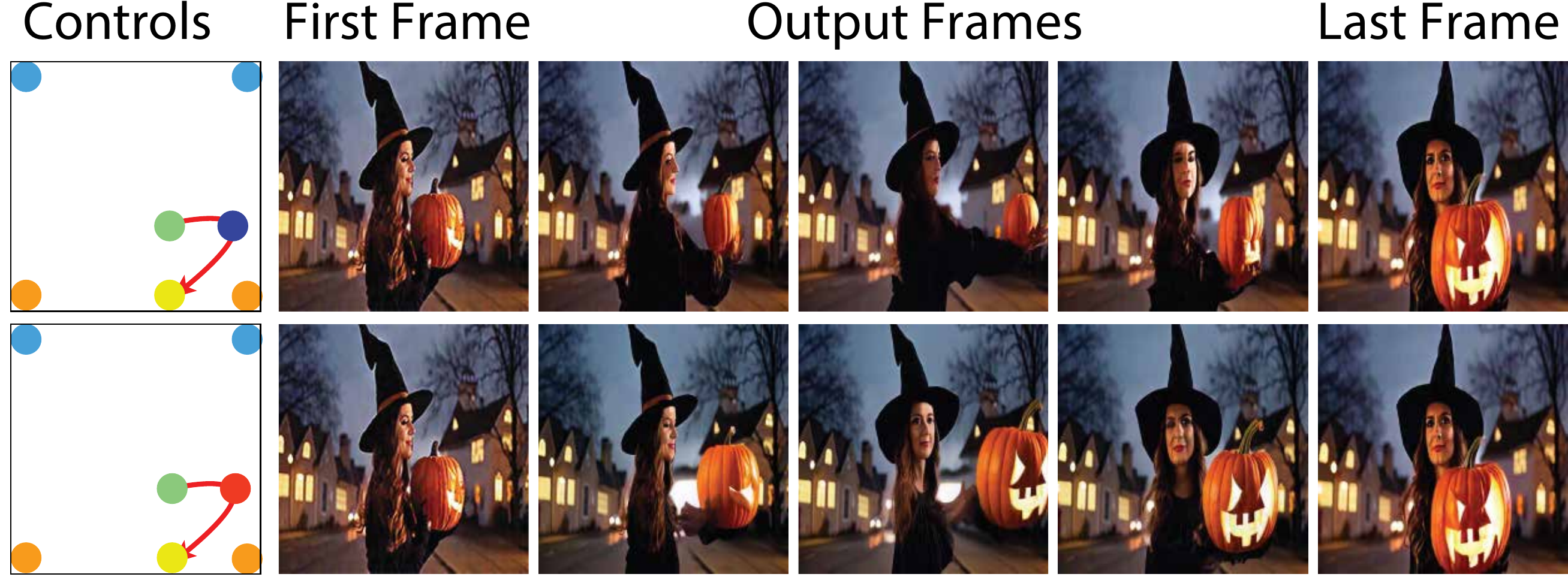}
  \caption{
  Example of a witch moving the Jack-o’-Lantern along the same path, with motion inward (top) or outward (bottom), depending on midpoint depth (blue vs. red dot).
  }
  \label{fig:sparse_depth}
\end{figure}

%% file: figs/main_results.tex
\begin{figure*}[!htbp]
\centering
  \includegraphics[width=1\linewidth]{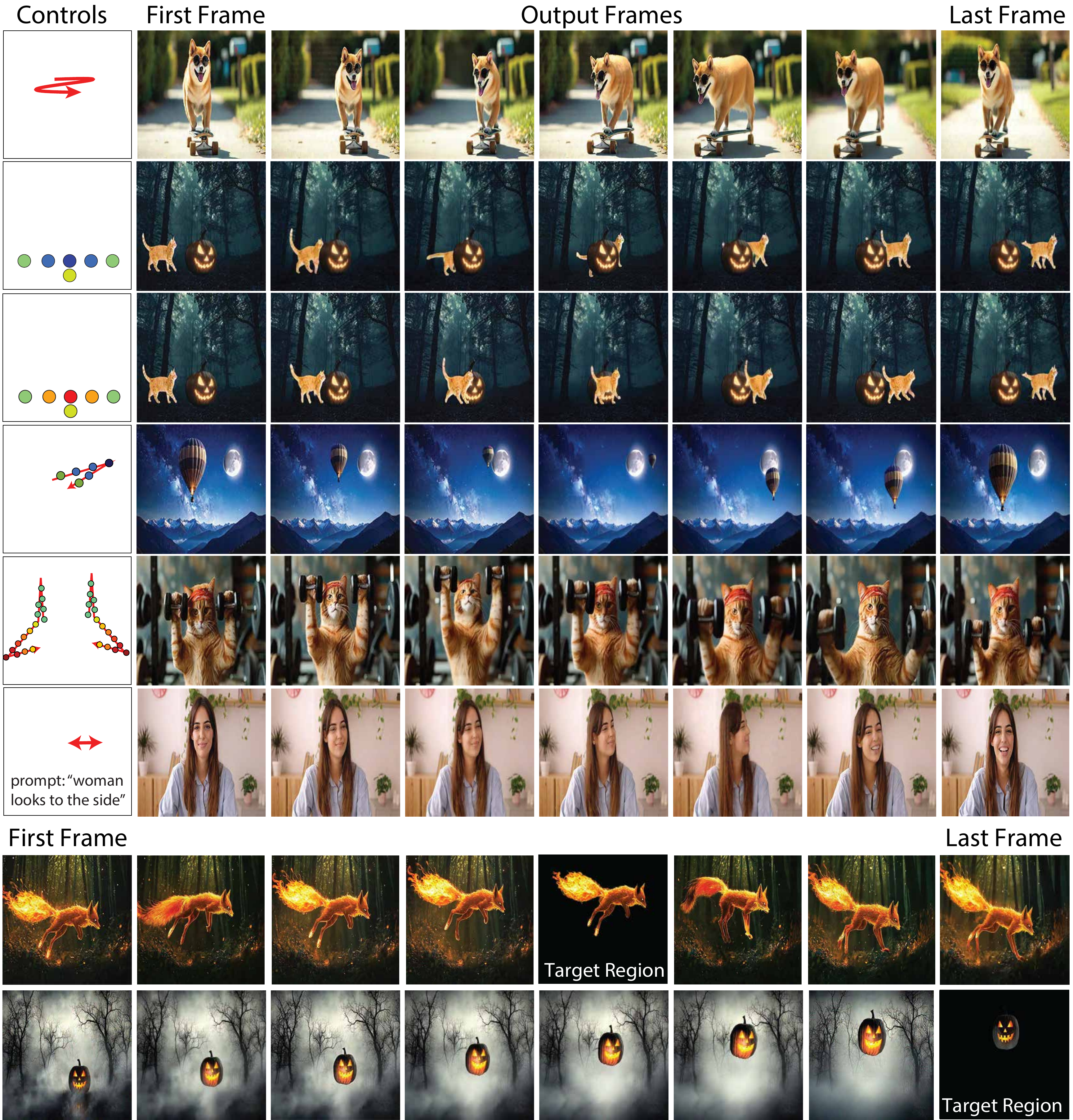}
  \caption{
Our results illustrate several ways multi-modal controls can be applied to frame interpolation. In the top section, we show trajectory control on its own, followed by two depth variations that place the cat either in front of or behind the pumpkin. Combining trajectory with depth produces richer motion: the balloon recedes along the z-axis while the weights with the cat are pushed outward. Prompts can also be paired with trajectories, where the trajectory sets the overall movement and the prompt refines details. In the bottom section, we highlight target region control. The temporal placement of target regions determines content editing at that point: in the first case, they are inserted in the middle with both first and last frames given, while in the second they appear at the end serving as a soft replacement for the last frame.
  }
  \label{fig:main_res}
\end{figure*}

% Our results demonstrate several ways the model can be applied creatively. The top row shows trajectory control. We then illustrate two cases with different depth settings: making the cat move in front of or behind the pumpkin by providing depth references. The next two rows show trajectory combined with depth control. Prompt can also be combined with trajectory for additional control. The final two rows present examples of using guide pixels.

% \begin{figure*}[t]
% \centering
%   \includegraphics[width=1\linewidth]{data/results_qualitative.pdf}
%   \caption{
%   \mt{Our results demonstrate several ways the model can be applied creatively. The top two rows show trajectory combined with depth control. Next, we illustrate two cases with different depth settings: making the cat move in front of or behind the pumpkin by providing depth references. Text can also be combined with prompts for additional control. A simple trajectory alone can be used as well. The final two rows present examples of guide pixels. In the first, we define a region of interest and move it along a trajectory to guide the model through regional control. In the last example, guide pixels are used purely for content control.}
%   }
%   \label{fig:main_res}
% \end{figure*}

%% file: sec/4_results.tex
\section{Experiments}
\label{sec:exps}
To evaluate \modelname{}'s performance, we conduct both quantitative and qualitative assessments across a range of video sequences and datasets. Currently, Framer~\cite{wang2024framer} is the only baseline that supports controllable interpolation, but it relies solely on trajectory control. Therefore, we compare our method to Framer under the trajectory control setting. For the quantitative evaluation, we assess both the generative quality and motion control of our model.

\input{figs/framer_comp}
\textbf{Implementation Details}:
We apply our method to a pretrained DiT text-to-video diffusion model, similar in architecture to OpenSora. The model is trained on the latent space of a 3D VAE that encodes 32 video frames into 5 latent frames. Training videos consist of 64 frames at resolution of 352x640, paired with text captions. The training uses 16 Nvidia A100 GPUs. We use an Adam optimizer with $1\times10^{-4}$ learning rate. Approximately 5k steps are used to train the image-to-video model, 2k steps for optical flow training, 2k for sparse and, 2k for target region input. The entire model, except the VAE and text encoders, is finetuned end-to-end.

\textit{Automatic Trajectory Generation}: 
For a fair quantitative comparison with baseline, we employ an automatic trajectory generation method similar to Framer ~\cite{wang2024framer}. Specifically, SIFT is used to extract features from the first and last frames of the video, and then point pairs are selected with high correspondence. A linear trajectory is generated between these matched points.

\textbf{Metrics and Datasets}: 
Following \cite{feng2024explorative, chen2023seine, wang2024framer} we use SSIM, FVD~\cite{unterthiner2019fvd}, and LPIPS~\cite{zhang2018unreasonable} for quality comparison.  Additionally, we introduce a ``Motion" metric to evaluate our model's trajectory control. This metric uses the optical flow of the generated output to create trajectory paths corresponding to the input trajectory, and we compute the Fréchet Distance to assess their similarity. 
We use DAVIS~\cite{pont20172017} and UCF (Sports Action)~\cite{rodriguez2008action} for analysis, as both feature large frame-to-frame motion across diverse cases.
\subsection{Qualitative Results}
\label{sec:exps_qual}
As shown in Fig.~\ref{fig:main_res}, our model integrates both content and motion controls, including trajectory, depth, target regions, and text. Trajectory produces smooth, realistic motion along the given path, while depth handles both relative cases (\eg a cat moving around a pumpkin) and single-point inputs (\eg a balloon). Combining trajectory and depth enables simultaneous 2D translation and depth variation. Text further refines outputs, and target regions provide intuitive content editing. The model also supports more than two input frames, as illustrated in Fig.~\ref{fig:mult_frame}, with/without motion control. Deformation is another application, for which results are shown in Fig.~\ref{fig:deform}.
\input{figs/multi_frame}
\input{figs/deform}
\subsection{Qualitative Evaluation}
\label{sec:exps_qual_eval}
We provide a qualitative comparison with the baseline Framer~\cite{wang2024framer} in Fig.~\ref{fig:framer}. We also attach videos for comparison in supplementary. Our model achieves smoother transitions with fewer distortions and artifacts, producing more natural interpolations. In Framer, motion is introduced as an external condition via ControlNet that interacts with video features indirectly. In contrast, our method embeds motion into the same latent space as the video, enabling stronger spatio-temporal alignment. This integration preserves frame quality while seamlessly incorporating user-defined motion and demonstrating the versatility of multi-modal control. In the second example, depth is leveraged to create compelling effects, while in the third, text serves as an additional condition to guide the model when trajectory alone is insufficient.
\input{tables/quant_interp}
\subsection{Quantitative Evaluation}
Quantitative results are reported in Tab.~\ref{tab:comparison_metrics} on the DAVIS~\cite{pont20172017} and UCF (Sports)~\cite{ahmadyan2021objectron} datasets. The motion metric demonstrates a clear improvement over the baseline in capturing motion trajectories. In terms of visual quality, our model generally matches or surpasses the baseline, with DAVIS-SSIM being the only exception, where we observe a minor decrease. Since SSIM is highly sensitive to pixel-level alignment, the slight drop in this metric is not critical. More importantly, the improved FVD indicates that our method produces perceptually more realistic and temporally consistent videos. Overall, our approach delivers comparable or superior visual quality with substantially stronger motion control.
\input{figs/abl_12}
\subsection{Ablation Study}
\textbf{The Effectiveness of Stage-wise Training.}
\label{sec:abl_opt}
We initially experimented with training directly on sparse motion and depth inputs. While this approach produced outputs with comparable perceptual quality, evidenced by FVD and LPIPS scores that remain on par with stage-wise training, the model failed to integrate the motion and depth cues effectively. As illustrated in Fig.~\ref{fig:abl_dual}, motion fails to localize accurately and depth information is misinterpreted. 

Quantitative results in Tab. ~\ref{tab:ablation} further confirm that motion control deteriorates severely without stage-wise training. Thus, even though the overall fidelity of generated frames is preserved, since the model was still trained on two-image interpolation and can hallucinate visually plausible outputs, the absence of stage-wise training leads to poor motion adherence, with the model failing to accurately follow the provided cues.
\input{tables/tab_ablations}
\noindent\textbf{The Effectiveness of Dual-branch Encoders.}
\label{sec:abl_twobranch}
In our system, content and motion are encoded through two dedicated branches. In this ablation, we replace the dual-branch design with a single branch, where all conditions are concatenated with the input noise. As shown in Fig.~\ref{fig:abl_dual}, this design leads to noticeably more artifacts when content and motion are not disentangled. The quantitative results in Tab. ~\ref{tab:ablation} further highlight a substantial decline in motion control, along with less realistic video quality as indicated by the higher FVD score.
\vspace{-5pt}

%% file: figs/framer_comp.tex
\begin{figure}[h]
\centering
  \includegraphics[width=1\linewidth]{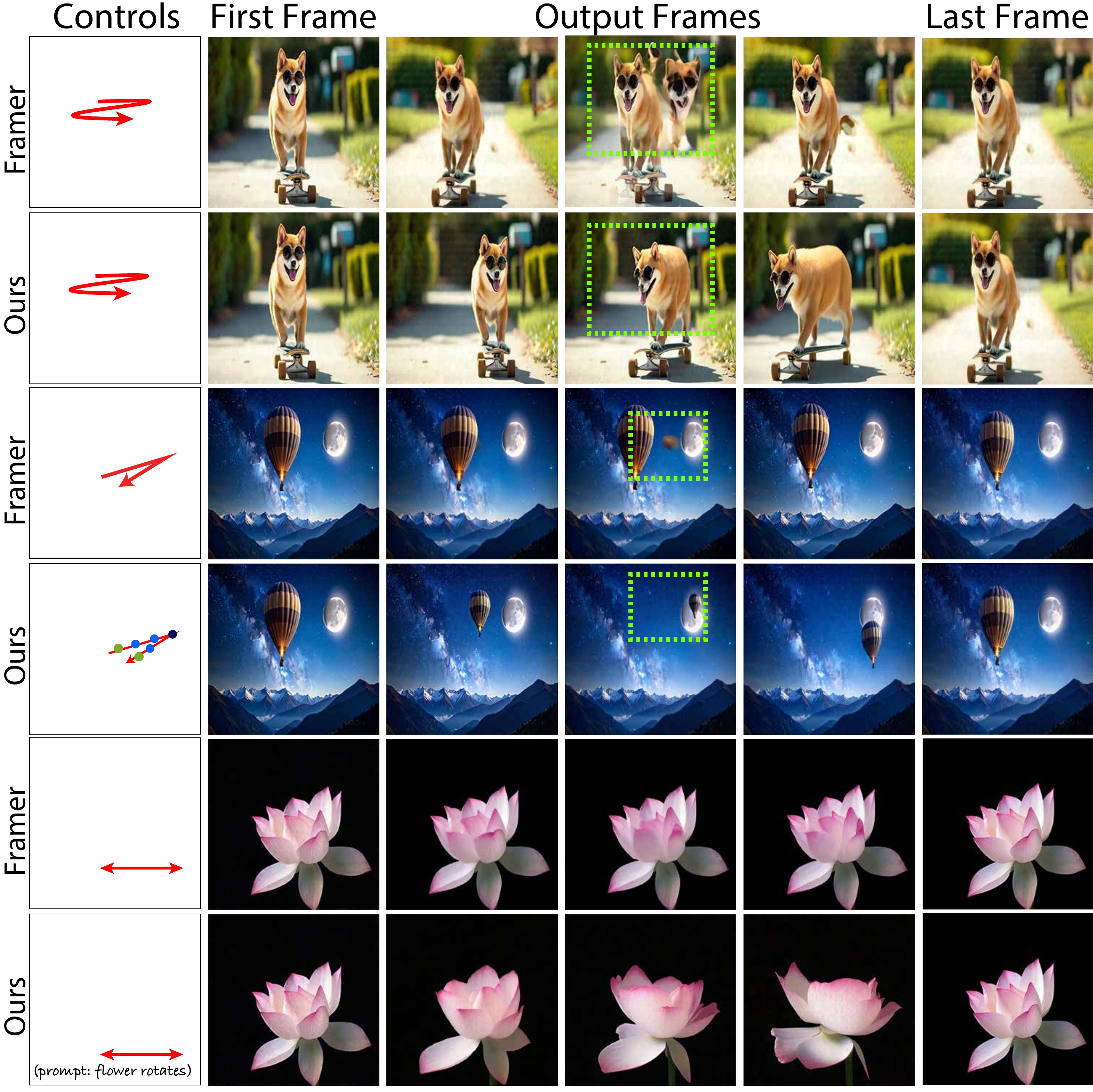}
  \caption{ 
  Comparison with Framer \cite{wang2024framer}. The top row highlights our reduced distortion for trajectory control, while the bottom rows showcase the benefits of additional controls such as depth control and text prompt.
  }
  \label{fig:framer}
\end{figure}

%% file: figs/multi_frame.tex
\begin{figure}[htbp]
\centering
  \includegraphics[width=1\linewidth]{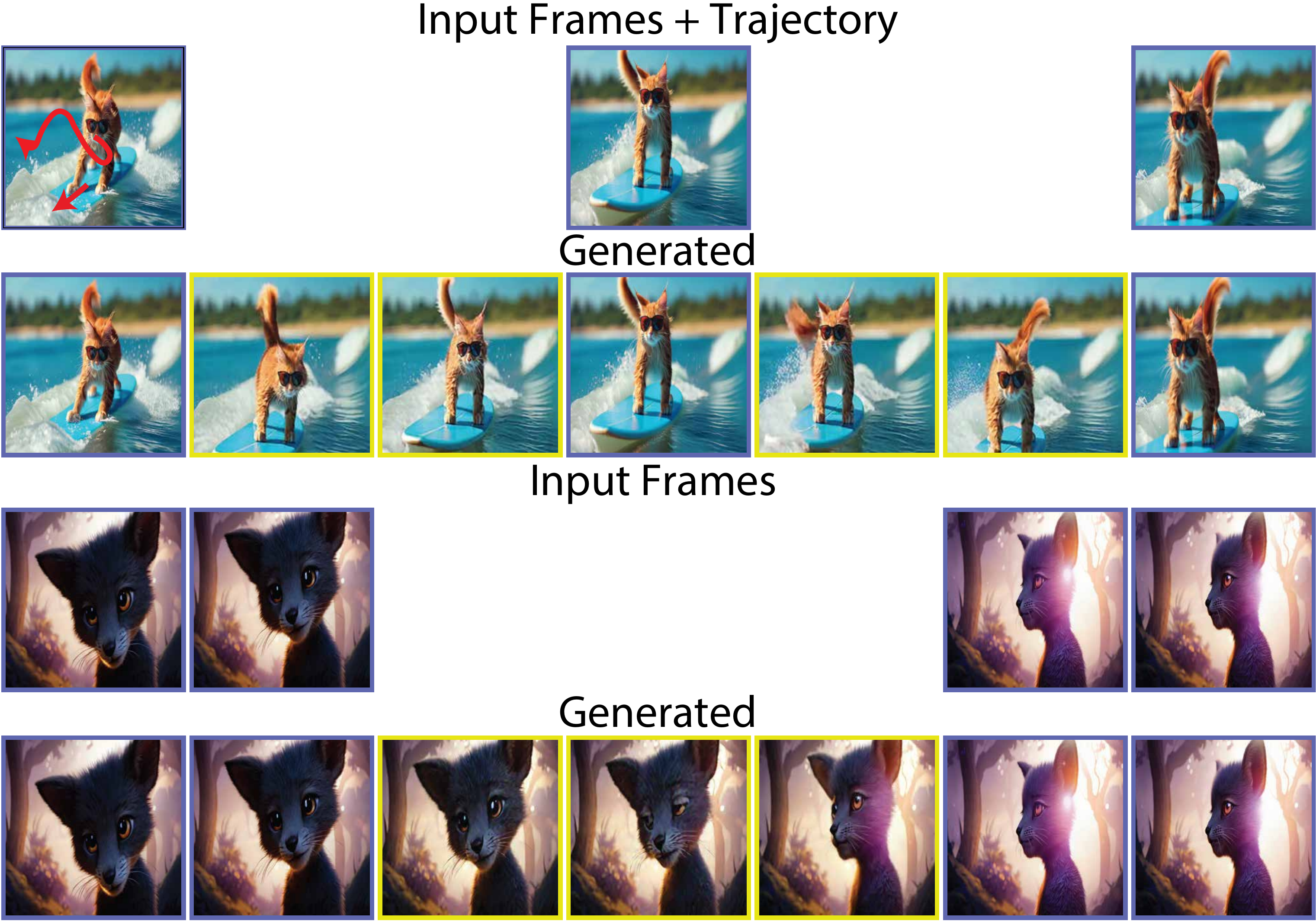}
  \caption{Results with more than 2 input frames, both with and without motion input.
  }
  \label{fig:mult_frame}
\end{figure}

%% file: figs/deform.tex
\begin{figure}[h]
\centering
  \includegraphics[width=1\linewidth]{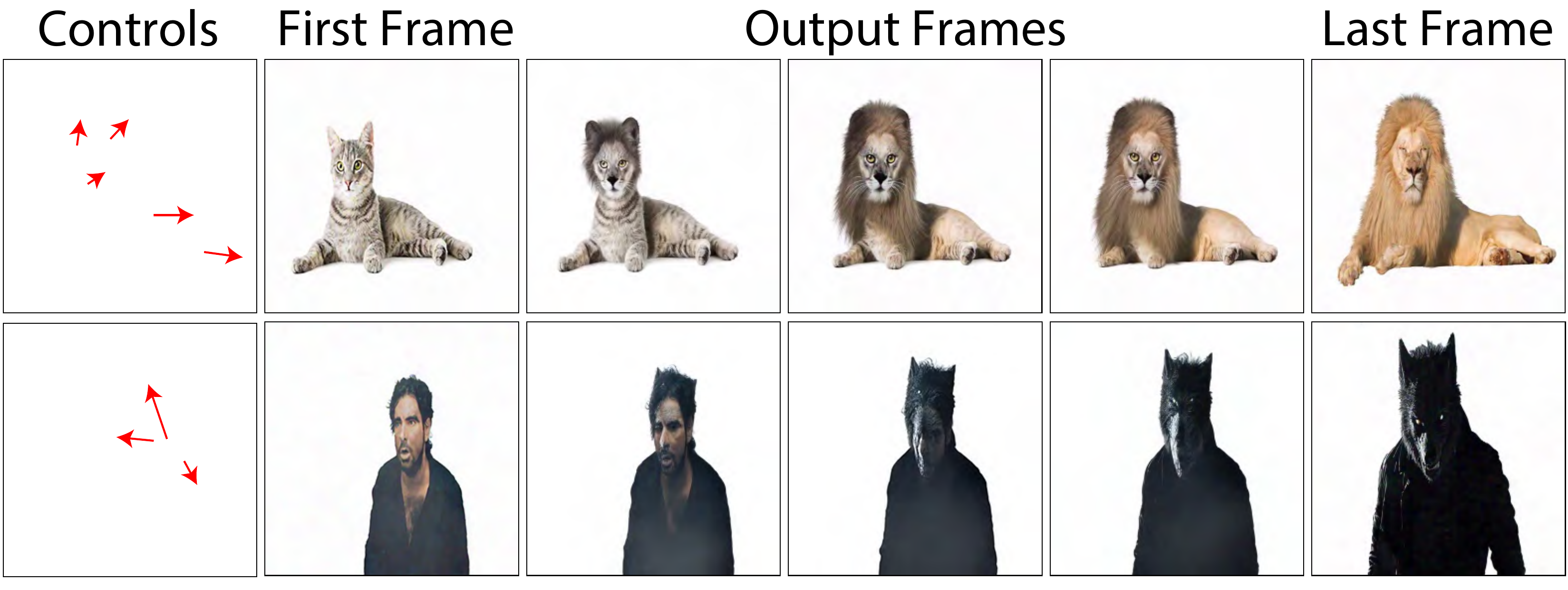}
  \caption{ 
  Results showcasing image deformation.}
  \label{fig:deform}
\end{figure}

%% file: tables/quant_interp.tex
% \begin{table}[h]
% \centering
% \begin{tabular}{lcccccccc}
% \toprule
% {\textbf{Model}} & \multicolumn{4}{c}{\textbf{DAVIS}} & \multicolumn{4}{c}{\textbf{UCF (Sports)}} \\
%                                 & FVD & LPIPS & SSIM & Motion & FVD & LPIPS & SSIM & Motion \\
% \midrule
% Framer                          & 4.42 & \textbf{0.50} & \textbf{0.18} & 5.25 & 2.14 & 0.48 & 0.21 & 3.31 \\
% Ours (no prompt)               & \textbf{4.33} & \textbf{0.50} & 0.16 & \textbf{2.44} & \textbf{2.58} & \textbf{0.31} & \textbf{0.34} & \textbf{2.34} \\
% \textit{Ours (+prompt)}        & \textit{4.30} & \textit{\textbf{0.51}} & \textit{0.15} & \textit{3.20} & \textit{2.44} & \textit{0.29} & \textit{0.37} & \textit{3.29} \\
% \bottomrule
% \end{tabular}
% \caption{Comparison of video generation models across two benchmarks. Lower is better for FVD, LPIPS, and Motion; higher is better for SSIM.}
% \label{tab:comparison_metrics}
% \end{table}

% \begin{table}[h]
% \centering
% \scriptsize
% \begin{tabular*}{\linewidth}{@{\extracolsep{\fill}}lcccccccc}
% \toprule
% \textbf{Model} & \multicolumn{4}{c}{\textbf{DAVIS}} & \multicolumn{4}{c}{\textbf{UCF (Sports)}} \\
%               & FVD ↓ & LPIPS ↓ & SSIM ↓ & Mot. ↓ & FVD ↓ & LPIPS ↓ & SSIM ↓ & Mot. ↓\\
% \midrule
% Framer           & 4.42 & \textbf{0.50} & \textbf{0.18} & 5.25 & \textbf{2.15} & 0.48 & 0.21 & 3.31 \\
% Ours & \textbf{4.33} & \textbf{0.50} & 0.16 & \textbf{2.44} & \textbf{2.14} & \textbf{0.31} & \textbf{0.34} & \textbf{2.34} \\
% \bottomrule
% \end{tabular*}
% \caption{Quantitative comparison with Framer.}
% \label{tab:comparison_metrics}
% \end{table}

\begin{table}[h]
\centering
\scriptsize
\setlength{\tabcolsep}{3pt} % default is 6pt, reduce for compactness
\begin{tabular*}{\linewidth}{@{\extracolsep{\fill}}lcccccccc}
\toprule
\textbf{Model} & \multicolumn{4}{c}{\textbf{DAVIS}} & \multicolumn{4}{c}{\textbf{UCF (Sports)}} \\
               & FVD↓ & LPIPS↓ & SSIM↑ & Motion↓ & FVD↓ & LPIPS↓ & SSIM↑ & Motion↓ \\
\midrule
Framer & 4.42 & \textbf{0.50} & \textbf{0.18} & 5.25 & \textbf{2.15} & 0.48 & 0.21 & 3.31 \\
Ours   & \textbf{4.33} & \textbf{0.50} & 0.16 & \textbf{2.44} & 2.14 & \textbf{0.31} & \textbf{0.34} & \textbf{2.34} \\
\bottomrule
\end{tabular*}
\caption{Quantitative comparison with Framer~\cite{wang2024framer}}
\label{tab:comparison_metrics}
\end{table}

% \begin{table}[h]
% \centering
% \scriptsize
% \setlength{\tabcolsep}{4pt} % adjust if needed
% \begin{tabular*}{\linewidth}{@{\extracolsep{\fill}}lcccc}
% \toprule
% \textbf{Model} & \multicolumn{4}{c}{\textbf{DAVIS}} \\
%                & FVD↓ & LPIPS↓ & SSIM↑ & Motion↓ \\
% \midrule
% Framer & 4.42 & \textbf{0.50} & \textbf{0.18} & 5.25 \\
% Ours   & \textbf{4.33} & \textbf{0.50} & 0.16 & \textbf{2.44} \\
% \midrule
% \textbf{Model} & \multicolumn{4}{c}{\textbf{UCF (Sports)}} \\
%                & FVD↓ & LPIPS↓ & SSIM↑ & Motion↓ \\
% \midrule
% Framer & \textbf{2.15} & 0.48 & 0.21 & 3.31 \\
% Ours   & \textbf{2.14} & \textbf{0.31} & \textbf{0.34} & \textbf{2.34} \\
% \bottomrule
% \end{tabular*}
% \caption{Quantitative comparison with Framer on DAVIS and UCF (Sports) datasets.}
% \label{tab:comparison_metrics}
% \end{table}

%% file: figs/abl_12.tex
\begin{figure}[h]
\centering
  \includegraphics[width=1\linewidth]{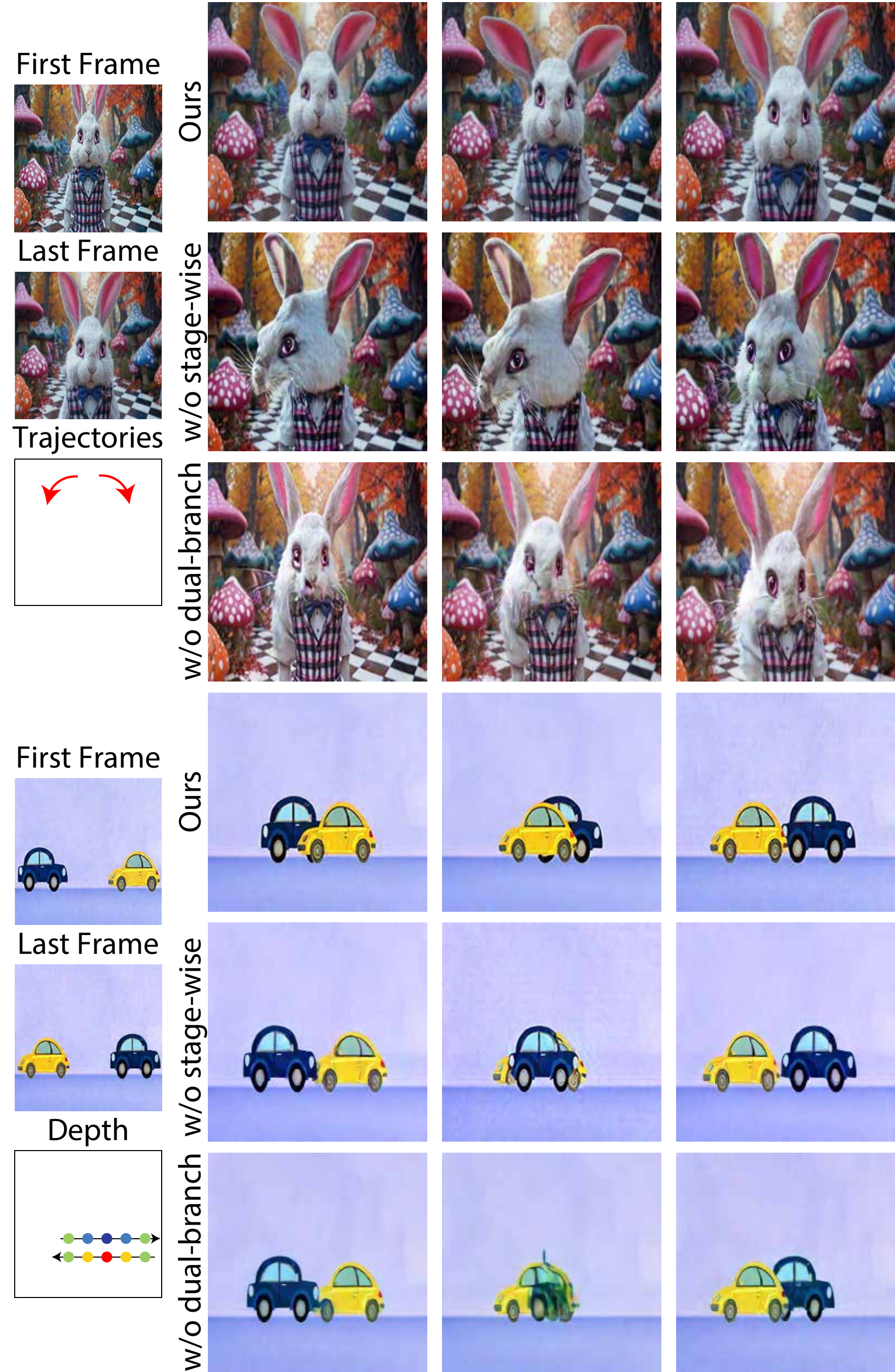}
  \caption{Ablation Results: w/o stage-wise: Skipping stage-wise training degrades performance; without dense flow the model mislocalizes motion, and without dense depth it misinterprets depth control.
w/o dual-branch: Removing the dual-branch design increases artifacts and causes depth confusion.
  }
  \label{fig:abl_dual}
\end{figure}

%% file: tables/tab_ablations.tex
% \begin{table}[H]
% \centering
% \resizebox{\columnwidth}{!}{%
% \begin{tabular}{lcccccc}
% \toprule
% & \multicolumn{3}{c}{\textbf{DAVIS}} & \multicolumn{3}{c}{\textbf{UCF (Sports)}} \\
% & FVD ↓ & LPIPS ↓ & Mot. ↓ & FVD ↓ & LPIPS ↓ & Mot. ↓ \\
% \midrule
% w/o stage-wise             & \textbf{4.25}  & \textbf{0.49}  & 4.41 & 2.32 & \textbf{0.26}  & 4.81 \\
% w/o dual-branch             & 5.90  & 0.53 & 3.24 &  3.28 & 0.32  & 4.93 \\
% \textbf{Ours}    & 4.33  & \textbf{0.50}  &  \textbf{2.44} & \textbf{2.14} & 0.31  & \textbf{2.34} \\
% \bottomrule
% \end{tabular}
% }
% \caption{Ablation study using w/o stage-wise and w/o dual-branch.}
% \label{tab:ablation}
% \end{table}

\begin{table}[h]
\centering
\scriptsize
\setlength{\tabcolsep}{1pt} % compactness
\begin{tabular}{lcccccccc}
\toprule
& \multicolumn{4}{c}{\textbf{DAVIS}} & \multicolumn{4}{c}{\textbf{UCF (Sports)}} \\
& FVD ↓ & LPIPS ↓ & SSIM ↑ & Mot. ↓ & FVD ↓ & LPIPS ↓ & SSIM ↑ & Mot. ↓ \\
\midrule
w/o stage-wise   & \textbf{4.25} & \textbf{0.49} & 0.14 & 4.41 & 2.32 & \textbf{0.26} & \textbf{0.38} & 4.81 \\
w/o dual-branch  & 5.90 & 0.53 & 0.13 & 3.24 & 3.28 & 0.32 & 0.33 & 4.93 \\
\textbf{Ours}    & 4.33 & 0.50 & \textbf{0.18} & \textbf{2.44} & \textbf{2.14} & 0.31 & 0.34 & \textbf{2.34} \\
\bottomrule
\end{tabular}
\caption{Ablation study using w/o stage-wise and w/o dual-branch.}
\label{tab:ablation}
\end{table}

%% file: sec/5_conc.tex
\section{Conclusion}
\label{sec:conc}
We introduced \modelname{}, a DiT-based framework for controllable inbetweening that generates high-quality interpolated frames conditioned on trajectories, depth, target regions, and text prompt. These conditions may be used individually or in combination with each other. Extensive experiments both qualitative and quantitative, demonstrate its versatility and effectiveness across a wide variety of use-cases. Nonetheless, challenges remain, particularly in aligning trajectories with image content, as strong content conditioning can dominate and suppress motion cues. Future work may incorporate lightweight pre-processing modules to better balance such controls, thereby preserving user intent while maintaining quality.